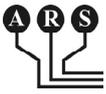

*Al-Jumail, A. & Leung, C. / Wavefront Propagation and Fuzzy Based Autonomous Navigation, pp.093-102, International Journal of Advanced Robotic Systems, Volume 2, Number 2 (2005), ISSN 1729-8806*# Wavefront Propagation and Fuzzy Based Autonomous Navigation

**Adel Al-Jumaily & Cindy Leung**
ARC Centre of Excellence in Autonomous Systems, Mechatronics and Intelligent Systems Group,
Faculty of Engineering,
University of Technology, Sydney
PO Box 123 Broadway NSW 2007 Australia
adel@eng.uts.edu.au, cleung@eng.uts.edu.au*Abstract: Path planning and obstacle avoidance are the two major issues in any navigation system. Wavefront propagation algorithm, as a good path planner, can be used to determine an optimal path. Obstacle avoidance can be achieved using possibility theory. Combining these two functions enable a robot to autonomously navigate to its destination. This paper presents the approach and results in implementing an autonomous navigation system for an indoor mobile robot. The system developed is based on a laser sensor used to retrieve data to update a two dimensional world model of the robot environment. Waypoints in the path are incorporated into the obstacle avoidance. Features such as ageing of objects and smooth motion planning are implemented to enhance efficiency and also to cater for dynamic environments.*
*Keywords: possibility theory, wavefront propagation, autonomous robot, indoor environment.*## 1. Introduction

Navigation is a major issue that needs to be addressed in order to facilitate the tasks required for autonomous systems. In particular, path planning and obstacle avoidance that enable an autonomous mobile agent to move effectively to its destination.

A principal challenge in the development of advanced autonomous systems is the realization of a real-time path planning and obstacle avoidance strategy which can effectively navigate and guide the vehicle in dynamic environments [Antonelli, 2001]. A path planned for an entity with partial knowledge of the environment can be invalidated with time. This can occur when unknown objects are detected or when objects move from their initial location, hence replanning of the path must be executed [Raulo, 2000].

The main purpose of the navigation system developed is to enable a mobile robot to navigate autonomously from its current location to any point on a map. In our work, the user need to provide the map. Specifications such as the size and resolution of the map are contained in the map provided. This map can be either empty (unexplored terrain with all cells unoccupied) or complete (contains walls). The update of the map with the objects, that are not included in the initial map, can be done by detecte such objects by the sensor. These objects are mapped and aged so they can be incorporated in the path and avoided effectively. Path planning is necessary to determine the gross motion required within the map and obstacle avoidance to modify the path to avoid collisions.

Navigation techniques of mobile robots are generally classified into reactive and deliberative techniques. The reactive technique is easily implemented by directly referring to sensor information. However robots may sometimes fall into a deadlock in complicated environments [Fujimori, 2002]. Deliberative techniques conversely use models such as environmental map for navigation. Robotic systems of this sort have the advantage of being able to produce an optimal plan from building complete maps, but they are limited in the usefulness due to lack of real-time reactivity to an uncertain or dynamic environment [Taliansky, 2000]. Deliberative planning and reactive control are equally important for robot navigation; when used appropriately, each compliments the other and compensates for the other's deficiencies [Rosenblatt, 1995]. It has proven useful for controlling mobile robots in man-made environments [Stoytchev, 2001].

The wavefront propagation algorithm is a deliberative technique since it finds the optimal path based on previously mapped information. Possibility theory on the other hand corresponds to a reactive technique due to its decision making conducted in real-time from sensor. By combining these two techniques, an optimal path can be planned using information from the

093

entire map and obstacles can be avoided by using laser readings within a range of 2R (i.e. twice the robot's width) minimising computation and hence increasing response to obstacles, allowing optimality and efficiency. Initial implementation and testing of this navigation system was conducted on a robot simulator and later ported to the physical robot. The following sections provide details of how the implementation of the system was approached. Section 2 provides details of the physical attributes and constraints of the robot. Section 3 explains the criteria used in selecting suitable path planning and obstacle avoidance algorithms. Section 4 provides an overview of the additional features implemented to enhance the performance of the system. Furthermore, an examination of the limitations and results obtained from experimentation are included in Section 5. Finally, Sections 6 and 7 contain the conclusion and recommendations for the future respectively.

**2. Physical Attributes**

This navigation system is developed for the Pioneer 2DX indoor mobile robot. It has a built in computer system and is equipped with a number of sensors such as sonar and lasers. The information from the laser is used for this project. This laser scans 180 degrees across the front of the robot.

In addition, the robot is equipped with position encoders allowing the displacement from the initial starting location to be calculated. This is used to localise the robot during simulation and testing.

Other attributes of the robot include differential motors enabling the robot to turn on the spot, and wireless communication allowing control from a remote computer. The robot and its devices are interfaced through a Player client. Simulation or the world is achieved using Stage. (Available from http://playerstage.sourceforge.net)

The dimensions of the robot are 33cm in width and 44cm in length. For this system, an assumption is made that the robot is round with a diameter 55cm which is the longest distance through the centre of the robot. By shrinking the robot conceptually to a single point, while the obstacle perimeter is enlarged by half of the robot's largest dimension allows the robot to be guided around obstacles. This method, known as "configuration space approach", is the easiest method to cater for the robot's dimensions. It works well with relatively small, circular-footprint mobile robots [Hong, 2000].

**3. Algorithms Implemented**

Several algorithms and techniques used to aid in autonomous navigation were analysed and assessed for their suitability for the implementation on an indoor mobile robot. The criteria used in the assessment were developed based on the scope and objective of the navigation system. The main objective is to enable the robot to move effectively to its destination. One of the assumptions made in the development of the system was that a two dimensional occupancy grid based map is provided by the user. Hence the algorithm selected is to be compatible with a grid based map. Information from the map is to be updated frequently according to new sensor information as a result the path needs to be replanned regularly. Taking in account that the algorithms requiring low amounts of computation are desirable to enable fast response to dynamic objects since the robot is continually processing sensor information. Two have been selected to be implemented for the navigation system.

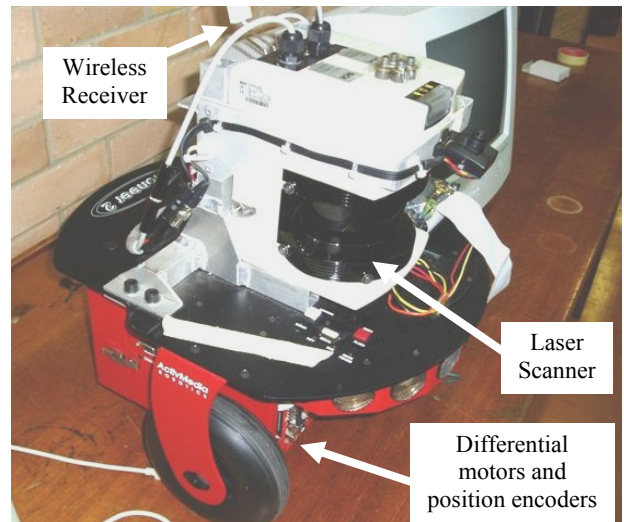

Fig. 1. Physical Attributes of Robot

*3.1 Wavefront Propagation Path Planning*
The wavefront propagation algorithm was chosen due to its suitability with grid based maps. This algorithm has emerged as the dominant method for path planning in discrete grid maps [Jennings, 1996]. The strategy is based on the propagation of wavefronts that encode the distance from the robot's current location to any point in its environment. As seen in Fig. 2, the wavefront propagation algorithm is applied to a simple grid based map. The wavefronts propagate from the source located on the centre right of the map. Each wavefront generated is designated a higher value than the previous. The shortest path can be determined by selecting any point on the map and then tracing the highest descent of wavefronts back to the source.

There are two main methods for computing the values of the wavefronts. The Manhattan style as demonstrated on the left in Fig. 3, only analyses adjacent cells in the grid to the cell in question, whereas the Chamfer method as seen on the right of the same Fig. 3, also computes cells on the diagonal. The Chamfer method is chosen to be implemented in this project as it yields a more direct path than the Manhattan style [Jennings, 1996]

Advantages of this algorithm are that it is simple, requires low computation, is able to find the shortest path, can deal with any shape object in the map and the resolution of the map does not significantly impede on the processing time required.

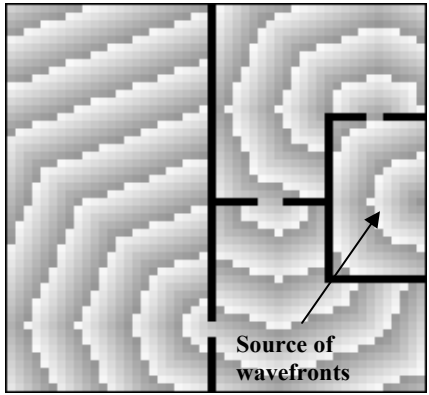

Fig. 2. Wavefront Propagation Applied to Map

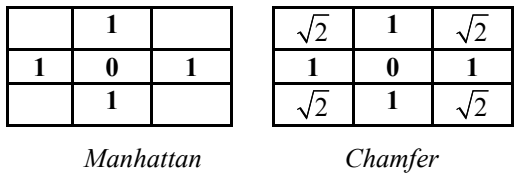

*Manhattan*         *Chamfer*

Fig. 3. Wavefront Propagation Methods

However there are some disadvantages with this algorithm that were to be overcome. These include: - assumption that the width of the robot is within a single grid cell and that the path planned cuts extremely close to walls or objects. Section 4 will describe the additional features implemented to overcome these limitations. In the event that the path becomes invalidated due to new sensor data, an obstacle avoidance algorithm is required to supplement this path planning algorithm.

*3.2 Possibility theory for Obstacle Avoidance*

When the robot detects that there is an obstacle in its path that it is following then it needs to be able react quickly to avoid it. Possibility theory algorithms, as a basis for fuzzy logic, were chosen to be implemented due to its ability to make decisions in real time and its ability to be tailored for the robot and the environment. It promises an efficient way for obstacle avoidance [Cang, 2003].

Possibility theory deals with the uncertainty [Dubois, 1996]; in this case the uncertainty of the location of obstacles. These uncertainties originate from the errors in the laser readings, errors due to lag as the robot is turning or moving at high speeds and noise from dust particles.

The rules of possibility theory are similar to probability theory, but use either MAX/MIN or MAX/TIMES calculus, rather than PLUS/TIMES of probability theory. PLUS/TIMES calculus however does not validly generalise nondeterministic processes, while MAX/MIN and MAX/TIMES do, giving it an advantage as a representation of non-determinism in systems [Drainkov, 2001].

As mentioned previously, only the objects on the map within an area with a radius of 2R of the robot are subjected to the possibility theory. This is to minimise computation and enhance response time to obstacles.

A possibility distribution is a normal fuzzy set where at least one membership grade equals one. Fig.s 4 and 5 show the membership functions of the normalised angle and distance fuzzy sets used to fuzzify the laser readings of each scan. These membership functions are then subjected to the fuzzy rules which were developed and adjusted based on experimental results.

The fuzzy rules in Table 1 and 2 are the reasoning used to determine the speed and turn-rate required of the robot based on the angle and distance of the obstacles. Negative turn-rates denote a clockwise turn and obstacles with a negative angle are on the right of the robot. Generally these rules, as shown in Table 1 state that if there is an obstacle on the right then turn left and vice versa. Table 2 states that if there is an obstacle close to the front of the robot then slow down.

Maximum of minimum fuzzy inference method was chosen to determine the speed and turn-rate fuzzy sets due to its simplicity. It is calculated for each reading in a laser scan within a range of 2R. Centre of Area was used in the implementation for defuzzification because it was deemed to be the ideal technique [Leyden, 1999]. This is calculated for the final fuzzy sets of the speed and turn-rate membership function as shown in Fig.s 6 and 7 respectively. To avoid hard-coding the rule when deciding which direction the robot should turn when there are obstacles directly in front of the robot, these rules are modified dynamically depending on the location of the next waypoint in the path. This is achieved by applying an additional condition to the rules in Table 2 for obstacles in the angle range of ZE. This condition checks the location of the next waypoint and determines if it is on the left or the right of the robot. The resultant rule would be to turn the robot towards the waypoint by choosing one of the alternatives shown in Table 2. Hence the path is incorporated in the obstacle avoidance decision making.

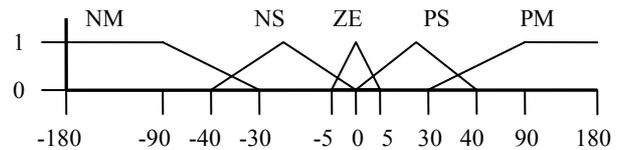

Fig. 4. Obstacle Angle Membership

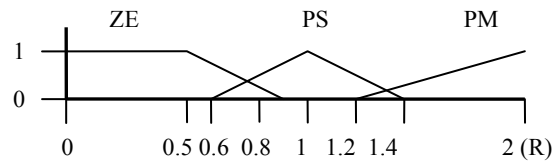

Fig. 5. Obstacle Distance Membership

| Speed    |    | Angle |    |    |    |    |
|----------|----|-------|----|----|----|----|
|          |    | PM    | PS | ZE | NS | NM |
| Distance | ZE | PS    | ZE | ZE | ZE | PS |
|          | PS | PS    | ZE | ZE | ZE | PS |
|          | PM | PM    | PS | ZE | PS | PM |

Table 1. Fuzzy Rules for Speed

| Turn-rate |    | Angle |    |          |    |    |
|-----------|----|-------|----|----------|----|----|
|           |    | PM    | PS | ZE       | NS | NM |
| Distance  | ZE | NS    | NM | NM or PM | PM | PS |
|           | PS | ZE    | NS | NM or PM | PS | ZE |
|           | PM | ZE    | NS | NS or PS | PS | ZE |

Table 2. Fuzzy rules for Turn-rate

The turn-rate is determined using this possibility theory algorithm. Once the turn-rate is computed to be close to zero then it is assumed that the obstacles have been avoided and a new path is planned. However this assumption is not always correct as there are occasions where the turn-rate may be computed to be zero while there are still obstacles around. This may occur when there are many obstacles surrounding both sides of the robot and the centre of area calculation may result in zero. This disadvantage of the possibility theory algorithm can be compensated by the wavefront propagation algorithm by re-planning and avoiding the obstacles. This is one example where these two algorithms complement each other in reducing their limitations and enhancing their advantages.

## 4. Additional Features of the System

Several features were implemented to compensate for the limitations of the algorithms chosen and to increase the safety of the robot. Other features were implemented to enhance the performance of the system in a dynamic environment and to provide a smooth motion.

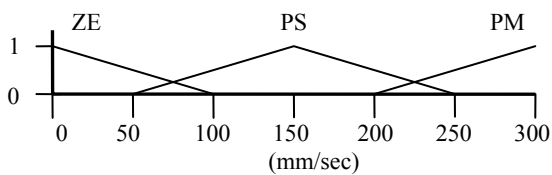

Fig. 6. Speed Membership

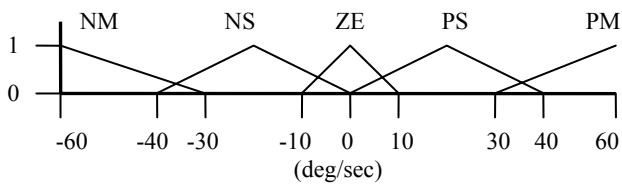

Fig. 7. Turn-rate Membership

*4.1. Thickening of Walls and Objects*

As mentioned previously in Section 2, the "configuration space approach" is used so obstacles need to be enlarged. The walls are thickened by half the robot's width and a designated safety distance. The thickening of walls solves two main issues encountered with the wavefront propagation algorithm. Since the wavefront propagation algorithm assumes that the robot is less than a single grid cell wide, paths can be planned though any gap in a wall that is a single grid cell. The other problem is that the path generated from this algorithm cuts extremely close corners to maintain the shortest path. In this case the path planned can be so close to the wall that it does not allow for the width of the robot or enough safety distance for the robot to pass. By thickening the walls, paths that are too narrow for the robot to pass are blocked and paths are planned further away from the actual wall. As a result, safer paths can be planned. Through experimental results a safety distance of 7cm and half the robot's radius (the robot is assumed to be round, so that there is always enough room for the robot to turn on the spot to escape from local minima) was determined to be the most beneficial by allowing sufficient space to manoeuvre and maintaining efficiency.

*4.2 Smooth Motion Planning and Waypoints*

Once the path has been planned then the robot is instructed to follow the path. However the path from the wavefront propagation algorithm is in the form of steps consisting of specific grid cells. It is difficult to follow a path cell by cell due to the accuracy and the constant stopping and checking for the next cell, hence the path is converted to a format that the robot can follow such as steps consisting of angle and distance to travel. This is achieved by determining the relationship between cells and grouping cells heading in the same direction.

Subsequent to this, there is still the problem that the path followed the lines of the wall. This means that if the wall is jagged then the resulting path would also be jagged. This problem is overcome by ignoring steps that were less than a designated minimum distance. The distance of the step ignored would be maintained in the path without the robot changing the direction. After this alteration to this path there is no longer a guarantee that the objects would be avoided. However with the assumption that the obstacle avoidance algorithm works in that it avoided any objects in the path that the robot takes then this minor detour from the initial path is allowable as it would increase the smoothness of the robot's motion quite significantly. The waypoints used for the obstacle avoidance algorithm is designated to be the last cell of each step of this smoothed path.

Once the jagged steps are removed, the robot continues to move in a stop-start motion as it must stop and turn before moving forward. This is not desirable as the motion does not seem natural or smooth. The solution implemented to solve this problem is to let the robot turn in an arc, much like how a car turns a corner as seen in Fig. 8. The larger the angle the robot is required to turn the smaller the radius of the arc. This increased the efficiency of the turn however it further increased the diversion from the initial path. This is due to the robot cutting corners as it turns an arc.

Hence there is increased reliance on the obstacle avoidance algorithm and the path planning is only used as a guideline on which path to take. Much like travelling in a car from one place to another, a path can be chosen by selecting certain roads to take. While the car is in motion, any obstacle avoidance or driving skill does not rely on the path chosen.

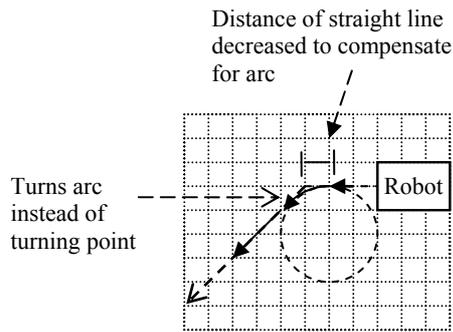

Fig. 8. Turns arc instead of turning on the spot

*4.3. Mapping and Ageing of Objects*
When moving around, it is desirable to ensure the data on the location of obstacles are correct. The map is updated with objects by mapping the laser readings. When the number of occurrences an object is detected to be in a particular cell exceeds a designated threshold the object is deemed to exist. In a dynamic environment objects can move or be moved. Hence new data is preferred [Singh 2000]. If all objects that were detected by the laser were mapped each time they were seen, and remapped as they changed their location, then the map could soon accumulate so many objects such that it would be completely occupied and there would be no more free space for the robot to move. The robot would be trying to avoid obstacles that are no longer in the same position as when it was detected.

Ageing of the objects allow for the objects that have not been recently detected by the laser to fade away as new and more recent information come to hand. This not only prevents the map from over cluttering with incorrect information but also removes the necessity for the system to repetitively perform ray-tracing to clear the grid cells between the robot and the obstacle detected. An effective ageing factor is determined from results of experimentation, variable depending on the speed of the robot and the threshold used for mapping. See 5.5 for resultant aging factor.

*4.4 Controlling Robot Movement*
The robot is controlled through the Player client and through the state machine that shown in Fig.9.
In Initialise State everything is initialised. Access is to the robot is requested and established. Access to all sensors established. The map file is obtained from the user, the current location of the robot on the map, and the destination is also obtained from the user. This state is entered upon start-up. It is left once all attributes have been initialised. While the prepare map state incorporates adding obstacles to the map and growing the obstacles of the map. This state is entered upon completion of initialisation procedures and is left once the map is prepared. It can also be entered from the obstacle avoidance state once the fuzzy logic controller has determined that the steering angle does not need to change. For the Plan Path the current attributes such as current location of the robot and the destination are checked to determine if a path can be planned. This state is left when the either the destination or the current location are occupied on the map. The path is planned in this state and once this is completed, it exits this state. If the path is not able to reach the destination this state is left.

While in the follow path, the robot is in motion and following the path. When the fuzzy logic controller determines that the current steering angle is required to change, the system leaves this state. This state is also exited when the destination is reached. At avoid obstacles state the robot follows the steering determined by the fuzzy logic controller. Once the steering angle no longer requires to be changed this state is left. This state is also left once the destination has been reached. Final state is stop state which in it the robot is stopped and the system is terminate.

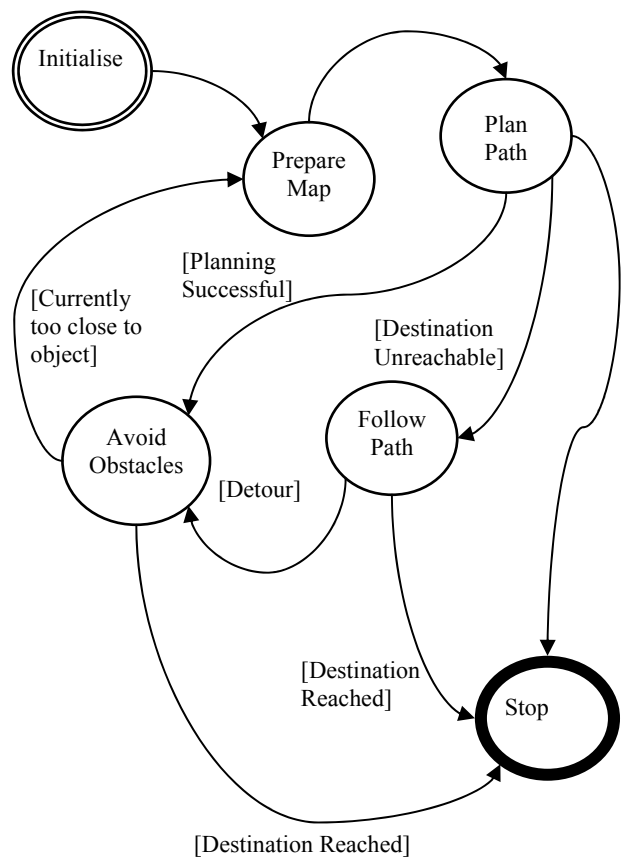

Fig. 9. State Diagram of robot movment

**5. Results of Experimentation**

Various functional and performance tests were conducted on the system. The system was tested under different scenarios to analyse how the system copes in different environments. Attributes of the system were also altered to determine how they affect the performance and functioning of the system.

*5.1 Limitation with Grids*
Function testing revealed several limitations of this system. This first limitation results from the limitation of dealing with occupancy grid based maps. Due to the layout of cells, each node processed in the path planning can only travel in eight directions limiting the angle of steps in a path to a minimum of 45°.

It is evident from Fig. 10 that with an empty map, the path still contains a turn. The shortest path between any two points is a straight line. Hence the path developed when the two points are not at a 45° angle is not the optimal path. This is deemed as an acceptable limitation because in reality most destinations cannot be reached by travelling in a straight line as the real environment contains many obstacles.

*5.2 Safety vs. Efficiency*

Another limitation discovered is that there are compromises between safety and efficiency. When there are walls for the path to be planned around, a safety distance is allocated and hence the path is no longer the shortest. It is evident from the first window in Fig. 11. that the resulting path from allowing the safety distance is not the shortest.

Issues that arose with a small safety distance include frequent alterations to the path triggered by the obstacle avoidance component. This component would detect the wall as an obstacle due to the small distance allowed for the robot to travel and hence an attempt would be made to move the robot away from this obstacle. By increasing the safety distance used in the path planning or decreasing the sensitivity of the obstacle avoidance then this interruption would be less frequent. However decreasing the sensitivity of the obstacle avoidance would increase the likelihood of a collision. Hence increasing the safety distance is the preferred option.

There are other concerns from increasing the safety distance in the path planning. When a path is to be planned through a narrow gap between walls, as seen in Fig. 11, the narrow path is completely blocked and hence the path is planned on a longer route. The path is further away from the wall hence the distance travelled also longer in this respect. Furthermore increasing the safety distance decreases the chance of a successful path planned. This is due to a grater chance that the robot lies in a grid cell that is occupied by the safety distance. Hence efficiency of the system is compromised when safety is taken into account.

Our soluation to solve these two problems is the thicke the walls to prevent the path being planned too close to the walls and it also fills in small gaps in the walls that are smaller than the width of the robot. Expanding obstacle boundaries to include the effective radius of a mobile robot allows the robot's center to be treated as a reference point.

Fig. 11 shows a path planned with a safety distance of 5cm. The walls contained in the initial map are in blue and the thickened walls are indicated by cyan. The path generated, in black, extends from the top right to the bottom right of the map, passing through the narrow gap. Due to the path's proximity to the wall there are many interruptions from the obstacle avoidance component as the robot follows this path. Fig. 12 shows a path planned with a safety distance of 15cm. This minimises the interruptions from the obstacle avoidance but the efficiency of the path is significantly diminished due to the blocked path. Determining a suitable safety distance is accomplished by analysing experimental results. A decision is made on how narrow a path can be before it is considered too narrow for the robot to traverse.

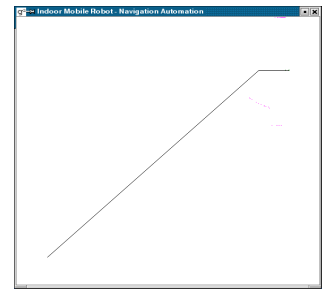

Fig. 10. Limitation to 45 degree turn in path

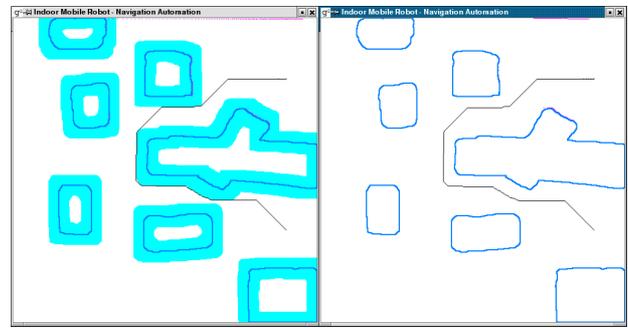

Fig. 11. Path planned with small safety distance

The amount of safety distance is required for moving at the set speed and how many interruptions from the obstacle avoidance are deemed acceptable for travelling a certain distance. As mentioned previously a safety distance of 7cm was deemed the most appropriate to cater for movement and maintaining efficiency. This safety distance allowed the physical robot to plan paths through doorways in the laboratory and also allowed the robot to manoeuvre through narrow gaps in the simulation world without too much interference from the obstacle avoidance component.

*5.3 Continuous Backtracking*

Further testing revealed other concerns when the robot would revert to a previously attempted path. This occurred under several scenarios. The first scenario is when the robot plans the shortest path to the destination and finds that this path is blocked. It then continues to plan another path which is longer than the first. As it starts following the second path the object is sufficiently aged so that when a path is re-planned, a path can be planned though the object having the shorter distance. The robot has yet to travel far enough to commit to the second path and hence returns to the blocked entrance. This scenario can continue forever. Decreasing the ageing factor of objects would reduce this problem by allowing more time for the robot to commit to the alternate route. However there are issues that become

098

apparent by decreasing the ageing factor. The main problem discovered is the maintaining of undesired information. For instance, errors are generated when the robot is turning due to lag. These errors result in the walls becoming thicker than they are in reality. Problems arise when narrow paths appear to be blocked due to this effect. Decreasing the ageing factor would maintain this error and thicken the existing walls preventing the traversal through what would otherwise be a possible shorter path. Hence when a path is replanned the longer route is selected.

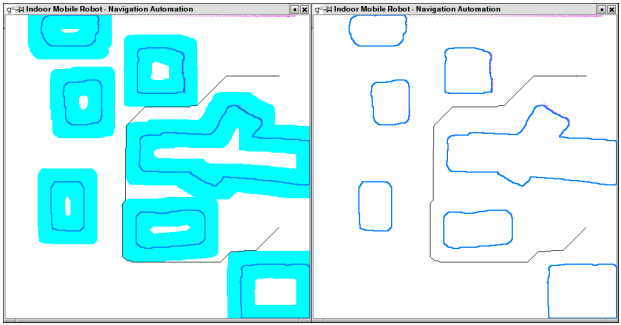

Fig. 12. Path planned with large safety distance

Alternatively if the ageing factor is high then this spray effect due to lag would not exist for long. Hence if the robot replans a path before taking the next step, the initial optimal path is regenerated and the robot would return to this narrow path. This may be a problem if the spiral effect continues to be regenerated as the robot would oscillate back and forth between the two paths.

Another scenario resulting in continuous backtracking is when the destination is completely surrounded by objects. The robot would travel towards the destination. Upon discovering that the path is blocked, it moves around to the other opening which is also found to be blocked. By this time the objects along the initial path would have aged sufficiently that a path can be planned though it, so the robot returns to the previous path.

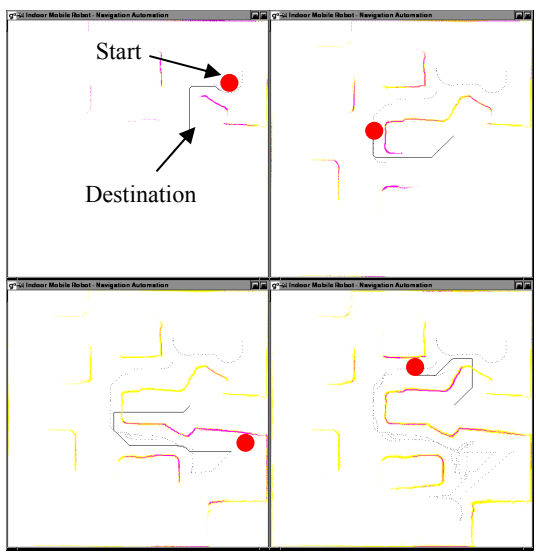

Fig. 13. Destination Unreachable

This can continue on forever until one of the objects is removed. As demonstrated in Fig. 13, the robot is provided with an empty map and the destination is completely surrounded. The recent objects are shown in magenta and the aged objects are in yellow. This scenario results from the ageing of static objects, primarily due to the assumption that objects can move or be moved. A solution to this problem is to enforce a timeout. If the robot takes a much longer time than expected to reach the destination then a timeout can be triggered. This requires estimation of travel time in regards to map size or path length. Alternately, determining which objects to age is another possible solution. If the value of an occupied cell is greater than a selected threshold then it may be regarded as static and hence not aged. This technique is considered to be viable. However due to time constraints, this navigation system was not retested with this feature. Although this system does distinguish between walls (initial data on the map provided) and objects whereby walls are not aged.

*5.4 Speed vs. Accuracy and Safety*

Increasing the speed of the robot does not always guarantee that the robot would reach its destination in a shorter time as other issues arise. The faster the robot moves, the less scans it takes of the same area. Therefore occupied cells have less opportunity to build up past the threshold and maintain the occupied state. This results in the same effect as a high ageing factor. Objects disappear quickly and the robot tends to return to a previously attempted path.

With greater speeds, the robot also requires stronger control action to avoid obstacles. Larger angles must be turned or reaction time must be decreased. This application is tailored for slower speeds of approximately 200mm/sec. If the robot is run at speeds higher than this limit then safety of the robot is compromised as the reaction time of the robot may not be quick enough or the angle turned may not be large enough.

Another issue discovered in testing is that walls appear jagged and to overlap when the robot is travelling at high speeds. This may be due to a greater difference between the forward speed and the turn-rate.

Furthermore, from running the robot at slower speeds, benefits result such as objects becoming well defined and having a lower impact if it crashes. However, it may take longer for the robot to reach its destination; hence safety and accuracy are compromised by speed.

The system performs well at the initial speed of 100mm/sec however this pace is too slow. Doubling the speed to 200mm/sec improved the efficiency of the robot and the obstacle avoidance was required to be adjusted to cater for this speed. When retested at 200mm/sec the system performed well. The speed was increased once more to 300mm/sec. This caused slight problems in the mapping due to increased lag as a result of the higher speed. Further increase of the speed made the system unpredictable. Hence testing on the physical robot was limited to a speed of 200mm/sec due to increased errors in the physical world while testing on simulation was allowed to be 300mm/sec due to an ideal environment.

*5.5 Effect of threshold on mapping*

As part of the component to avoid obstacles, objects are mapped so they can be processed in relation to the location of the robot. When mapping objects, the system takes scans of the environment at different instances of time and accumulates this information. A threshold is applied to the mapping of objects to filter out noise and undesired information.

Increase of this threshold results in a faster rate of ageing of objects. Static objects in the environment take longer to be detected and dynamic objects would struggle to appear on the map. Dynamic objects are aged significantly faster than static objects. This is due to the cell count of the map having a lower value because the object is not detected to be in the same cell for long periods of time. Objects appear and disappear quickly with a high threshold. Hence system response to avoid obstacles is greatly diminished. Similarly to applying a high ageing factor, the limitation of undesired backtracking is also apparent with a high threshold.

Decrease of the threshold results in objects appearing faster and are maintained longer. However errors in readings such as the readings taken when the robot is turning that give a spraying effect also appear. Hence the system's response incorporates obstacles that exist and those that are errors in readings. Consequently the system becomes extremely sensitive and prone to error.

Therefore the threshold used to map objects has similar limitations and effects as the ageing factor. System response is compromised by obtaining valid data. A threshold of 7 was found to be the most suitable for this system with an ageing factor of about 0.14 with a speed of 300mm/sec in simulation. Whereas an threshold of 10 with an aging factor of 0.1 with a speed of 200mm/sec performed well in the real-world environment.

*5.6 Empty Map vs. Complete Map*

The direction of movement and planning is based on the map provided and the initial location of the robot positioned on the map. There were many differences found between using an empty map and a complete map. This was only tested in simulation due to the unavailability of a complete map of the walls of the laboratory. The obvious difference is that the initial path planned in an empty map could be far from the final path taken whereas the initial path planned on a complete map remains fairly close to the path travelled. Hence a complete map would result in a more efficient path.

Another obvious difference, apparent as a result of an inadequate localiser, is that the starting location does not have to be specific with an empty map as the robot can be positioned anywhere. However when using a complete map, the robot must be positioned fairly accurately on the map so as the objects sensed are closely related to those on the map otherwise errors would accumulate significantly. These issues would be solved pending integration with a decent localiser which determines the location of the robot based on data retrieved from the environment.

*5.7 Exploration*

When the robot is provided with an empty map exploration can be conducted. The system treats all the walls as objects hence anything detected by the laser is mapped. Once the objects are mapped they can be avoided and hence the robot is able to reach its destination without much prior knowledge of the environment. It only requires the general direction in which to travel in relation to its current position denoted by the destination provided by the user. This was achieved successfully both on the physical robot and in simulation. However at times, the robot may run off the map when avoiding obstacles because it is of the fixed map size provided by the user.

Fig. 14 shows an exploration sequence that was conducted during testing. The first window shows an initial path generated from an empty map from the robot's location to the destination. This path is a straight line because there were no obstacles on the map when it was generated. As the robot starts following this path it discovers that there are obstacles in the way which invalidates the current path. When these obstacles are avoided, the system generates a new path around the obstacles as seen in the window on the top right. This step is repeated as the robot continues to move and discover new objects. The old objects start to age as depicted by the fading pixels in the third window.

Continuing on to the fourth window, the centre right window in Fig. 14, the objects that were first detected have aged sufficiently that the robot is able to plan a path through them thinking that it is a shorter path and the objects have moved. It returns to discover that the objects are still there and plans a new path once again. The robot follows this new path and continues to map objects and regenerate the path until it reaches the destination as demonstrated in the final two widows.

This test proved the robot is able to successfully reach its destination while avoiding the objects.

*5.8 Dynamic Objects*

In one other scenario that was tested, the robot is placed in a world with two other robots. These two other robots move about in a random manner while avoiding walls. The robot is to move from the top right of the map to the bottom left of the map, as seen from the initial path planned in the top right window of Fig. 15 The sequence of windows in Fig. 15 show the robots moving in the world on the left and current information gathered about the world on the right.

The first set of windows in Fig. 15 shows the robot following the initial path. It then discovers that there is an obstacle in its path and avoids it by applying possibility theory on the laser readings even before the object been detected a sufficient amount of times to be second set of windows. The robot then follows the new path until the destination is reached. The path remain relatively smooth except when the system forces the mapped. Once avoided and the object is mapped, a new path is planned avoiding the obstacle, as seen in the robot to follow a new path after obstacles are avoided in which case the path may be heading a completely different direction to the initial path as seen in Fig 14. A similar scenario is created for the physical robot. The two random walking robots are replaced with humans standing in the robot's path. The robot uses the same algorithm to successfully navigate around the humans.

*5.9 Map resolution*

The system was tested with a map with a resolution of 500 by 500 cells and a 100 by 100 cell map, both representing an area of 10m by 10m, as shown in Fig. 16. There was no noticeable difference in the speed of processing map information for path planning and obstacle avoidance although it is obvious that a larger map would require larger amounts of processing. Both map sizes generated paths in a few milliseconds. Perhaps a larger variation in the resolutions would provide a more apparent difference.

robot's current position. The robot is then instructed to move to its destination while avoiding obstacles and treating walls as objects. This was achieved successfully however the majority of experimentations were conducted in simulation due to the short battery life of the Pioneer.

# 6. Conclusion

This paper has discussed the design and results of experimentation of the navigation system developed for an indoor mobile robot. In conclusion, the goal of enabling the mobile robot to effectively reach its destination through autonomous navigation is achieved successfully. This approach has proven to work under various scenarios.

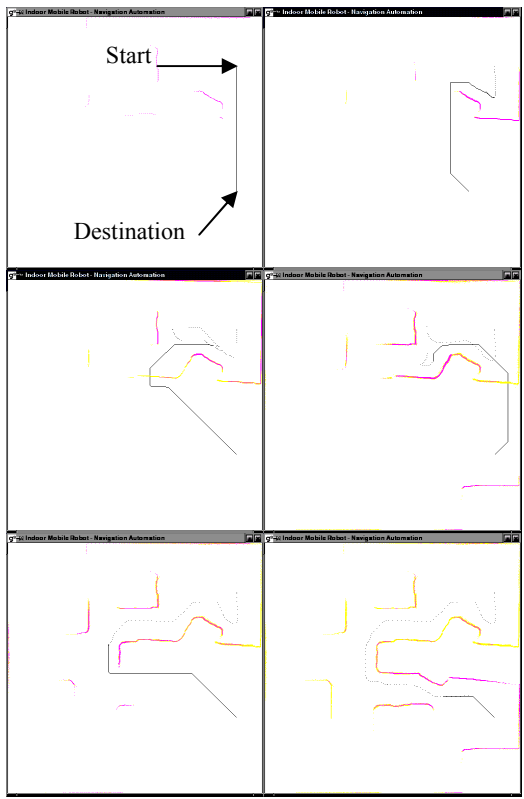

Fig. 14. Exploration Sequence

The main noticeable difference between these two sizes was the detail contained in the map. The map with the higher resolution is more accurate and contains more detail than the lower resolution map. Therefore there is a compromise between accuracy and speed. Accuracy is preferred because this is not a direct relation. Although, having a map with a higher resolution than the precision of the laser readings or position proxy is useless.

*5.10 Simulation vs. Real-world*

Experimentation on the physical robot was conducted using an empty map, due to the unavailability of a map of the laboratory. The robot is placed at an arbitrary position at the centre of the map facing 90 degrees. The destination is the selected as a distance relative to the

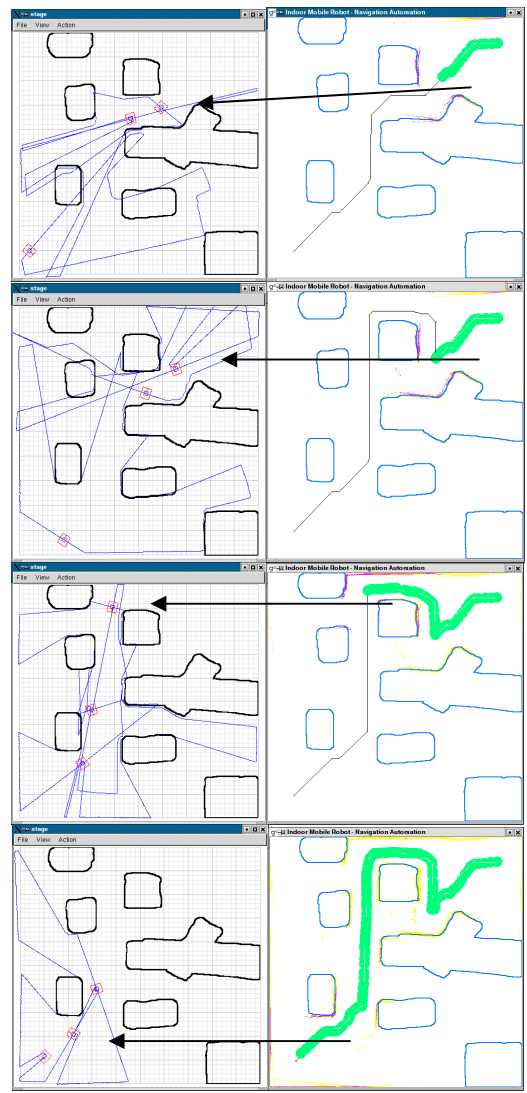

Fig. 15. Dynamic Objects

By incorporating the waypoints into the obstacle avoidance only minor deviations from the path were taken while avoiding obstacles. Obstacles that are not somewhat symmetrically surrounding the robot are easily avoided, however when the robot approaches a dead-end a new path is required to be generated to avoid the obstacles. By allowing a 7cm safety distance around the robot, the path remains fairly optimal. It is determined that the resulting path taken by the robot is relatively safe and efficient.

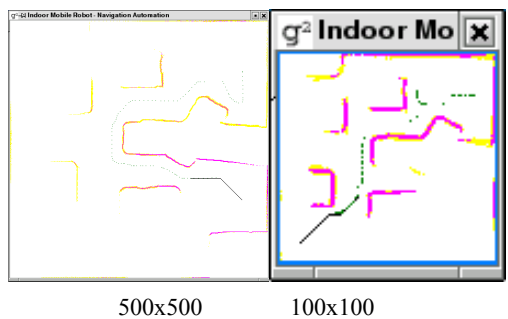

500x500     100x100

Fig. 16. Map resolution

### 7. Future Recommendations

This navigation system can be improved if it can be integrated with a localiser that determines the robot's position according to information obtained from the environment. A method such as a particle filter can be used. The result of this would minimise accumulated errors and reduce the impact of any errors in the initial location of the robot obtained from the user.

Although this system is not focussed on map building, there were occasions when the robot ran off the map. When this occurs the system would stop the robot and the user would need to restart the system placing the robot back onto the map. This system can be extended to enable the map to be expanded as the robot moves or estimate the location of the robot and move it back onto the map.

A method to be developed that determines the difference between static objects and dynamic objects would resolve the many limitations causing the robot to continuously backtrack, as found during performance testing. Not ageing objects that have a cell value greater than a designated threshold could be implemented as a viable solution. Or possibly ageing the static objects with a lower ageing factor could prove to be a better solution. Further experimentation is required to determine the legitimacy of this solution.

Paths are not always direct due to the limitation of format of grid cells. A method to relax the path could be implemented so that the robot is able to travel through paths that are not so rigid.

### Acknowledgements

This work is supported by the ARC Centre of Excellence programme, funded by the Australian Research Council (ARC) and the New South Wales State Government.